\documentclass[runningheads]{llncs}

\usepackage[T1]{fontenc}
\usepackage{graphicx}
\usepackage{caption}
\usepackage{subcaption}
\usepackage{array}
\usepackage{float}
\usepackage{amsmath}
\usepackage{booktabs}
\usepackage[table,xcdraw]{xcolor}
\usepackage{placeins}
\usepackage{listings}
\usepackage{xcolor}
\usepackage{placeins}  
\usepackage{hyperref}

\definecolor{amethyst}{rgb}{0.6, 0.4, 0.8}
\definecolor{darkpastelgreen}{rgb}{0.01, 0.75, 0.24}
\definecolor{amber}{rgb}{1.0, 0.75, 0.0}
\definecolor{cadmiumorange}{rgb}{0.93, 0.53, 0.18}
\definecolor{lawngreen}{rgb}{0.49, 0.99, 0.0}
\definecolor{limegreen}{rgb}{0.2, 0.8, 0.2}
\definecolor{neongreen}{rgb}{0.22, 0.88, 0.08}
\definecolor{amethyst}{rgb}{0.6, 0.4, 0.8}
\definecolor{darkpastelgreen}{rgb}{0.01, 0.75, 0.24}
\definecolor{greenbest}{RGB}{88,137,15}
\definecolor{redworst}{RGB}{137,15,27}
\definecolor{royalazure}{rgb}{0.25, 0.41, 0.88}

\lstdefinelanguage{json}{
     basicstyle=\ttfamily,
     showstringspaces=false,
     breaklines=true,
     frame=single,
     literate=
      *{0}{{{\color{blue}0}}}{1}
       {1}{{{\color{blue}1}}}{1}
       {2}{{{\color{blue}2}}}{1}
       {3}{{{\color{blue}3}}}{1}
       {4}{{{\color{blue}4}}}{1}
       {5}{{{\color{blue}5}}}{1}
       {6}{{{\color{blue}6}}}{1}
       {7}{{{\color{blue}7}}}{1}
       {8}{{{\color{blue}8}}}{1}
       {9}{{{\color{blue}9}}}{1}
       {:}{{{\color{red}:}}}{1}
       {,}{{{\color{red},}}}{1}
       {\{}{{{\color{orange}\{}}}{1}
       {\}}{{{\color{orange}\}}}}{1}
   }

\begin{document}
\title{BodyShapeGPT: SMPL Body Shape Manipulation with LLMs}
\author{Baldomero R. Árbol\orcidID{0000-0002-8320-9752} \and
Dan Casas\orcidID{0000-0002-3664-089X}}
\authorrunning{B. R. Árbol and D. Casas}
%
\institute{Universidad Rey Juan Carlos, Madrid, Spain\\
\vspace{0.3cm}\href{https://github.com/baldoarbol/BodyShapeGPT}{https://github.com/baldoarbol/BodyShapeGPT}}
%

\maketitle              
\begin{abstract}
Generative AI models provide a wide range of tools capable of performing complex tasks in a fraction of the time it would take a human. Among these, Large Language Models (LLMs) stand out for their ability to generate diverse texts, from literary narratives to specialized responses in different fields of knowledge. This paper explores the use of fine-tuned LLMs to identify physical descriptions of people, and subsequently create accurate representations of avatars using the SMPL-X model by inferring shape parameters. We demonstrate that LLMs can be trained to understand and manipulate the shape space of SMPL, allowing the control of 3D human shapes through natural language. This approach promises to improve human-machine interaction and opens new avenues for customization and simulation in virtual environments.

\end{abstract}
\section{Introduction}
The increasing integration of artificial intelligence in our daily lives challenges us to continually explore and expand its capabilities to enhance human-machine interaction. Large Language Models (LLMs) represent a revolution in how machines understand and generate human language, offering virtually unlimited possibilities in terms of content creation and response automation. This research aims to leverage the capabilities of LLMs to interpret complex descriptions and convert them into accurate visual representations, using the SMPL-X~\cite{Bogo:ECCV:2016,SMPL-X:2019} model to generate three-dimensional avatars from textual descriptions.
Current works~\cite{lucas2022posegptquantizationbased3dhuman,feng2024chatpose,jiang2024motiongpt} that utilize LLMs to modify SMPL parameters primarily focus on adjusting the pose rather than the shape. 
Additionally, while some methods~\cite{zeng2023avatarboothhighqualitycustomizable3d,cao2024dreamavatar} generate avatars from text, they do not provide explicit control of SMPL shape space.
Our approach addresses this gap by enabling the control of SMPL shape parameters directly from textual descriptions, enabling the interactive manipulation and generation of human avatars using natural language.

\section{Related Work}
\subsubsection{Avatars from Image.}In the field of computer vision, generating 3D avatars from images is a widely explored research area~\cite{alldieck2018detailed,yang2024havefunhumanavatarreconstruction,casas2023smplitex,casasEG2014,jiang2022instantavatar}.
SHAPY~\cite{Shapy:CVPR:2022} utilizes a deep learning architecture to analyze the photograph and reconstruct a detailed 3D model, capturing the nuances of human shape and appearance.
POCO~\cite{dwivedi_3dv2023_poco} focuses on achieving high-fidelity shape generation by leveraging convolutional neural networks to extract and translate image features into a 3D form.
AG3D~\cite{dong2023ag3d} synthesizes high-quality 3D humans from 2D image collections, capturing detailed appearances and geometries, including various clothing styles. 
AvatarBooth~\cite{zeng2023avatarboothhighqualitycustomizable3d} introduces a novel method for generating high-quality personalized 3D avatars using casual face or body images, and it also supports generation and editing based on text. FitMe~\cite{Lattas_2023_CVPR} presents a facial reflectance model and a differentiable rendering optimization pipeline that can capture detailed human avatars from one or multiple images, achieving photorealistic results and preserving identity.
StyleAvatar3D~\cite{zhang2023styleavatar3d} uses image and text diffusion models to generate high-quality stylized 3D avatars, while DreamAvatar~\cite{cao2024dreamavatar} employs a combination of textual and shape guidance to produce controllable 3D human avatars with specific poses.

This paper differs in that it does not require a photograph as input. Instead, the only input to the network is a \textit{textual description of the body shape} of a human subject.
This allows for greater flexibility and application in fields such as storytelling and virtual character creation, where body descriptions are common and often preferred.

\vspace{-0.1cm}
\sloppy
\subsubsection{Animations from Text.}In the realm of text-based generative networks, PoseGPT~\cite{lucas2022posegptquantizationbased3dhuman}, MotionGPT~\cite{jiang2024motiongpt} and ChatPose~\cite{feng2024chatpose}, among others~\cite{delmas2022posescript,dabral2022mofusion}, are notable for their ability to interpret text inputs and transform them into avatar poses or brief movement sequences. PoseGPT~\cite{lucas2022posegptquantizationbased3dhuman} leverages a transformer-based architecture to decode textual descriptions into precise body poses, which can then be used in various animation and virtual reality applications. MotionGPT~\cite{jiang2024motiongpt} extends this capability by generating continuous motion sequences from text, allowing for dynamic and expressive avatar animations. ChatPose~\cite{feng2024chatpose} further advances this field by integrating conversational context to generate more nuanced and contextually appropriate avatar movements.

Our work is based on a similar approach but with the goal of generating the avatar shape instead of the pose.
By focusing on shape generation, this work complements pose generation methods \cite{lucas2022posegptquantizationbased3dhuman,jiang2024motiongpt,feng2024chatpose}, offering a solution for creating complete avatars from natural language.

\section{Methodology}
\subsection{Dataset Generation}
Since there is no existing dataset that relates SMPL-X shape parameters to their corresponding verbal descriptions, a new dataset has been meticulously created to train the network for this task.
The dataset generation process began by taking measurements of randomly sampled avatars.
These measurements were then analyzed to understand their distribution within a random set of avatars with a normal distribution of body shapes.

In addition to these measurements, a custom labeling algorithm was developed to assign detailed verbal descriptions to each avatar.
Our strategy took into account various aspects of the avatar physical attributes, ensuring that each verbal description was comprehensive and precise.
For example, an avatar with a certain shoulder width and waist size might be described as having a "broad-shouldered and slim-waisted" appearance.
By carefully correlating these measurements with descriptive language, the labeling process generated a significant volume of data.

After the initial creation of the dataset descriptions, an additional layer of variability was introduced by rephrasing the text using Llama-3~\cite{dubey2024llama3herdmodels} to rewrite the same information with different wording and attribute order, enhancing the richness of the dataset.

The combination of detailed measurement analysis, precise verbal labeling, and varied rephrasing ensures that the neural network can learn the complex relationships between physical shapes and their descriptive counterparts, leading to more accurate and realistic avatar generation based on text input.
Figure \ref{fig:json_example} shows two samples of our dataset.

   \begin{figure}[h]
     \centering
     \begin{lstlisting}[language=json, gobble=0]
     {"description": "Person with an average height, tall neck, long arms, and broad shoulders.", 
     "shape_params": "[1.131, 1.928, -2.347, -0.793, 0.251, 0.58, 1.707, -2.888, -1.904, 2.772]"}
     {"description": "Medium-built individual with a slender profile, featuring a mid-range stature, a petite torso, and limbs of moderate length.", 
     "shape_params": "[-1.016, -0.504, 0.948, 1.092, -0.514, -1.941, 0.415, 2.089, 0.509, 1.626]"}
     \end{lstlisting}
     \caption{Example of dataset entries that relate SMPL-X shape parameters to their corresponding verbal descriptions}
     \label{fig:json_example}
   \end{figure}

\vspace{-1cm}
\subsection{Fine-Tuning Process}
Our model is trained by fine-tuning the popular Llama-3~\cite{dubey2024llama3herdmodels} model to the specific task of interpreting verbal descriptions of avatars and generating the corresponding SMPL-X parameters.
We use Low-Rank Adaptation (LoRA)~\cite{hu2021loralowrankadaptationlarge} and quantization~\cite{dettmers2023qloraefficientfinetuningquantized} to optimize the model for efficient training and execution on NVidia RTX-4090.
The fine-tuning was carried out starting from LLaMA-3 8B, using the generated dataset consisting of 18,000 training cases and 2,000 evaluation cases. 
To train our model, we propose to extend the regular Cross Entropy loss used to train LLMs. 
Our key intuition is that the output of the model is not arbitrary text, but a string of characters that must encode a vector of 10 float values (i.e., the SMPL shape parameters).
This observation allows us to formulate two new shape-specific loss terms that enable the model to focus on shape parameters rather than arbitrary text. 
Our loss term is defined as
\begin{equation}
    \mathcal{L} = \mathcal{L}_\text{LLM} + \mathcal{L}_\text{shape} + \mathcal{L}_\text{measurements}
\end{equation}
%
%
where $\mathcal{L}_\text{LLM} = \text{CE}(\hat{Y}_t, Y_t)$ is the standard cross-entropy loss of tokens used to train LLMs, which ensures that the output is a correct string of digits.
The term $\mathcal{L}_\text{shape} = |\hat{\beta} - \beta|$ is the L1 difference between shape parameters, weighting each beta according to its impact on the human parametric model.
Finally, $\mathcal{L}_\text{measurements} = \text{CE}(f_\text{SMPL}(\hat{Y}_t), f_\text{SMPL}(Y_t))$, where $f_\text{SMPL}$ is the SMPL function that evaluates a body mesh given $Y$ shape parameters,  is the cross-entropy loss of a set of body measurements (e.g., height, arm length, volume).
This term is key to guarantee that the generated avatar falls within the expected range.

\vspace{-0.3cm}
\section{Results}
\subsection{Quantitative Evaluation}
%

To validate our loss function, we carry out an ablation study for each of the terms.
To this end, we evaluate the following models
\begin{itemize}
    \item $\mathcal{L}=\mathcal{L}_\text{LLM}$: only considers the cross-entropy error of the generated text tokens.
    \item $\mathcal{L}=\mathcal{L}_\text{LLM}+\mathcal{L}_\text{shape}$: takes into account both the cross-entropy error of the generated text tokens and the error the shape parameters (i.e. $\beta$ parameter of SMPL).
    \item $\mathcal{L}=\mathcal{L}_\text{LLM}+\mathcal{L}_\text{shape}+\mathcal{L}_\text{measurements}$: our full model, it considers the cross-entroy of the text tokens, the error in shape parameters, and and also adds the cross-entropy of a set of body measurements, including limb lengths, BMI, and height.
\end{itemize}

\begin{figure}
\includegraphics[width=\textwidth]{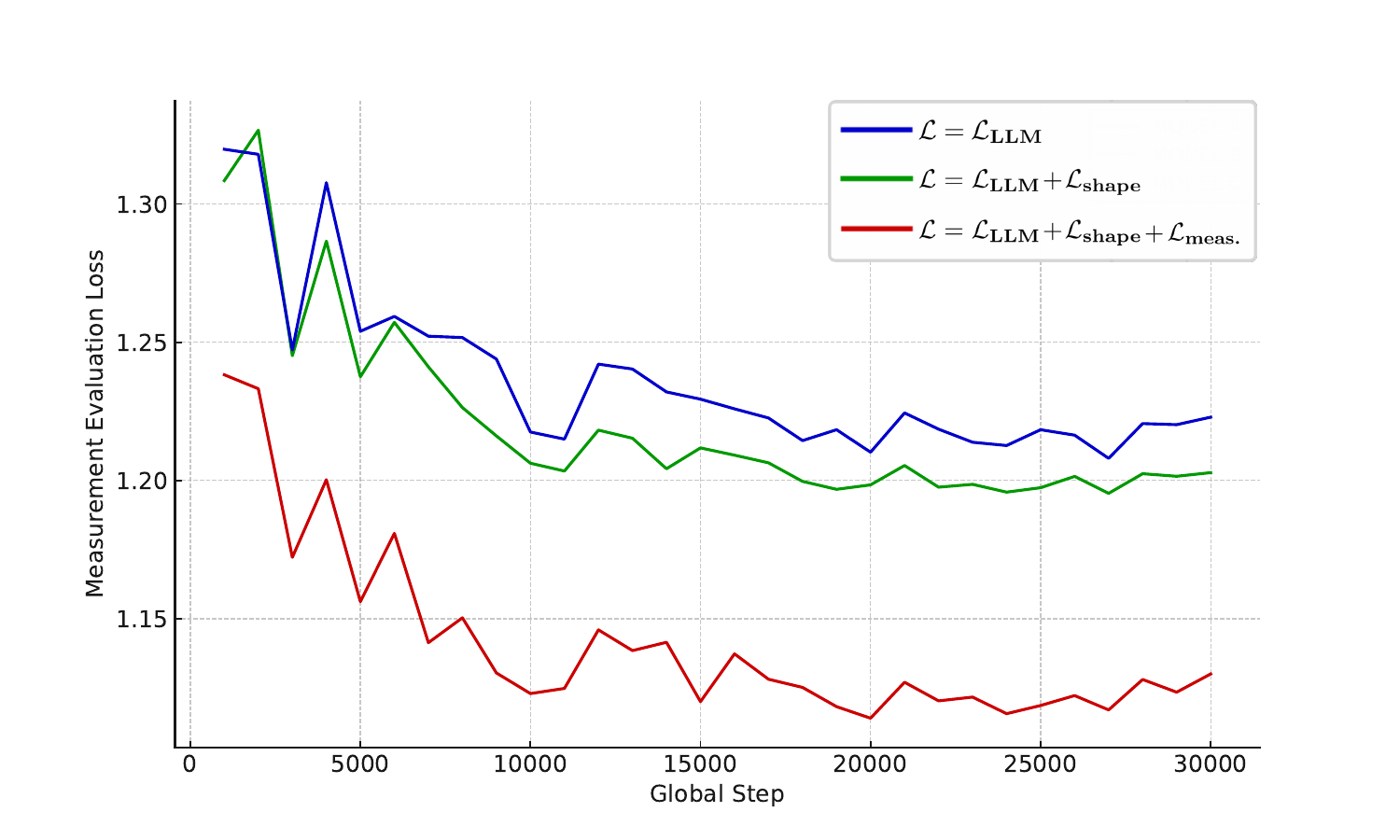}
\caption{Cross Entropy Loss of the validation set for each training iteration.} \label{graph_measurement_loss}
\end{figure}

As we show in Figure \ref{graph_measurement_loss}, our proposed loss achieves lower error than other baselines.
The error depicted in the graph represents the difference between the ranges in which the avatars are labeled and the desired target ranges.
A lower error indicates that the model predictions are more accurate in matching the desired labeling ranges, leading to more precise and consistent avatar descriptions.

The significant reduction in the cross-entropy loss value for the various measurements represents a clear improvement in the model's results regarding the accuracy of body measurements within the desired ranges. Studying its reduction throughout the training process helps identify the impact of the applied improvements.



\begin{figure}
  \centering
    \begin{subfigure}[b]{0.48\textwidth}
         \centering
         \includegraphics[width=\textwidth,trim={10pt 0 30pt 0}]{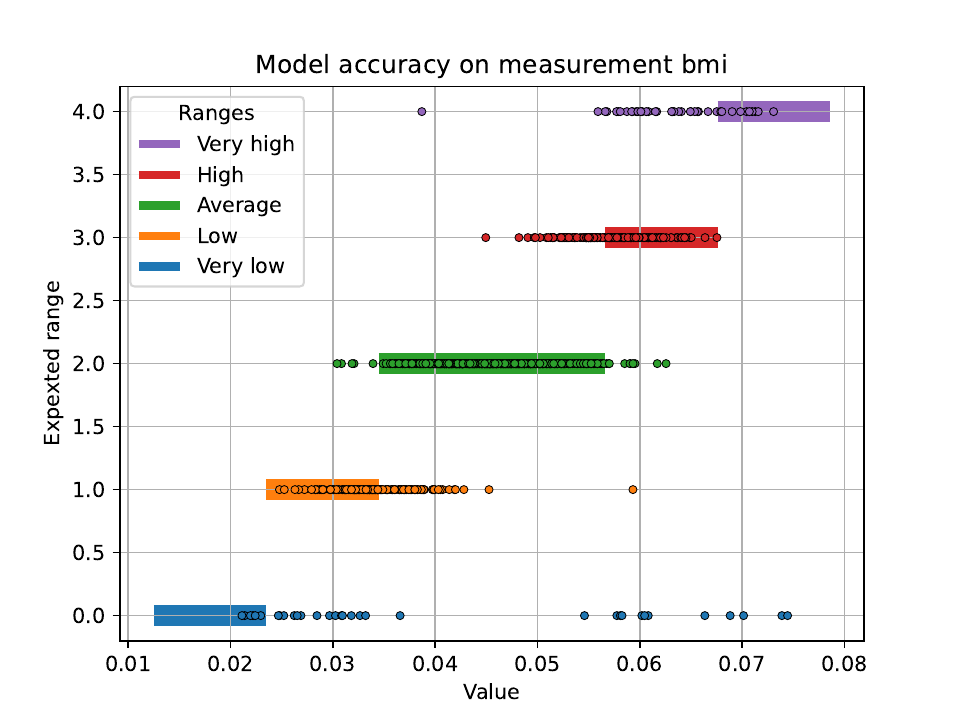}
         \caption{$\mathcal{L}=\mathcal{L}_\text{LLM}$}
         \label{accuracy_modelA}
     \end{subfigure}
     \hfill
     \begin{subfigure}[b]{0.48\textwidth}
         \centering
         \includegraphics[width=\textwidth,trim={10pt 0 30pt 0}]{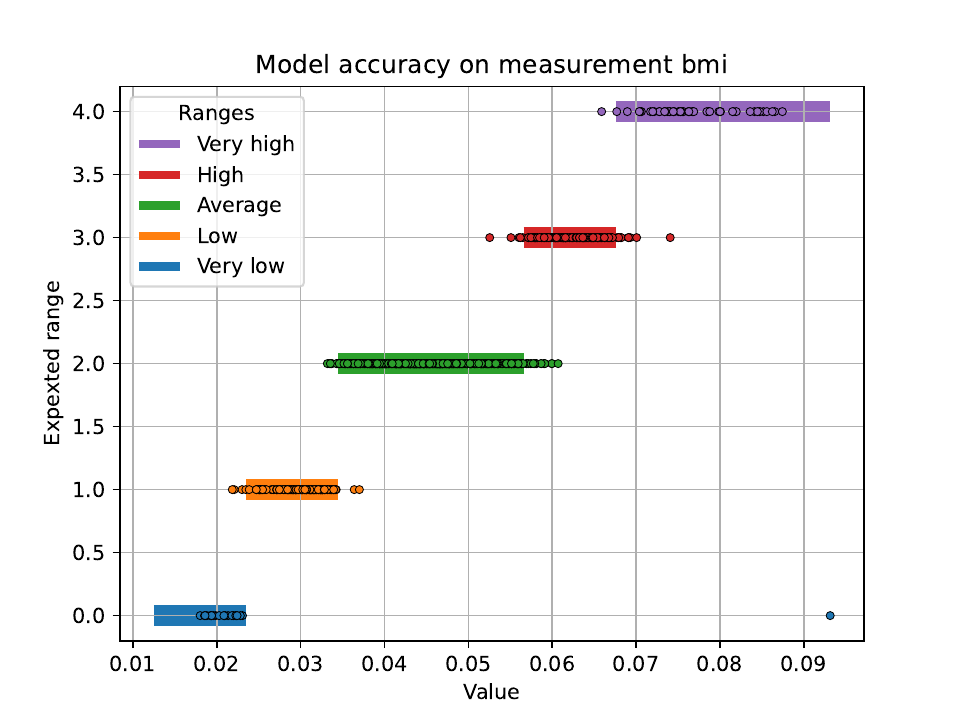}
         \caption{$\mathcal{L}=\mathcal{L}_\text{LLM}+\mathcal{L}_\text{shape}+\mathcal{L}_\text{meas.}$}
         \label{accuracy_modelC}
     \end{subfigure}
  
  \caption{Accuracy of baseline model (a) and our model (b), evaluated using the Body Mass Index (BMI) computed from the regressed avatars. The ground truth range for each category is shown as horizontal solid boxes. Using our model, most of test samples (i.e., color dots) fall into the correct range. }
  \label{fig:accuracy}
\end{figure}

This accuracy can also be observed in the plots from Figure \ref{fig:accuracy}, which evaluates the
accuracy of the distributions of our predictions.
It is worth noting that the accuracy of baseline model, which considers fewer factors when calculating error, is much lower, tending to generate avatars with measurements very close to the average and making errors typical of only considering text tokens as the value to account for, such as representing "very low" values as "very high." Our model, on the other hand, has a distribution where the values fit much better with the ground truth and align with the descriptions within the expected range.

Table \ref{tab:results} depicts the numerical accuracy of each of the models. Our model achieves the highest accuracy on all the measurements that we evaluate.

\begin{table}
\caption{Measurements accuracy using different models.}\label{tab1}
\centering
\begin{tabular}{|l|c|c|c|}
\hline
\rowcolor[HTML]{F4CCCC} 
 & \textbf{$\mathcal{L}_\text{LLM}$} & \textbf{$\mathcal{L}_\text{LLM}+\mathcal{L}_\text{shape}$} & \textbf{$\mathcal{L}_\text{LLM}+\mathcal{L}_\text{shape}+\mathcal{L}_\text{meas.}$} \\ \hline
\rowcolor[HTML]{FFFFFF} 
\textbf{height\_average} & 99,7\% & 99,1\% & \textbf{99,9\%} \\ \hline
\rowcolor[HTML]{E6E6E6} 
\textbf{hip\_thickness\_average} & 95,4\% & 96,1\% & \textbf{99,1\%} \\ \hline
\rowcolor[HTML]{FFFFFF} 
\textbf{hip\_thickness\_average} & 94,8\% & 96,0\% & \textbf{98,8\%} \\ \hline
\rowcolor[HTML]{E6E6E6} 
\textbf{bmi\_average} & 94,5\% & 95,8\% & \textbf{98,4\%} \\ \hline
\rowcolor[HTML]{FFFFFF} 
\textbf{waist\_thickness\_average} & 94,5\% & 94,9\% & \textbf{98,0\%} \\ \hline
\multicolumn{4}{|c|}{\dots} \\ \hline
\rowcolor[HTML]{FFFFFF} 
\textbf{neck\_length\_very\_high} & 33,3\% & 50,0\% & \textbf{82,6\%} \\ \hline
\rowcolor[HTML]{E6E6E6} 
\textbf{arms\_relation\_very\_high} & 30,4\% & 41,6\% & \textbf{81,0\%} \\ \hline
\rowcolor[HTML]{FFFFFF} 
\textbf{leg\_thickness\_very\_low} & 24,0\% & 41,6\% & \textbf{79,3\%} \\ \hline
\rowcolor[HTML]{E6E6E6} 
\textbf{shoulders\_relation\_very\_low} & 21,7\% & 41,1\% & \textbf{78,1\%} \\ \hline
\rowcolor[HTML]{FFFFFF} 
\textbf{bmi\_very\_low} & 18,7\% & 40,9\% & \textbf{76,7\%} \\ \hline
\end{tabular}
\label{tab:results}
\end{table}

\vspace{-0.3cm}
\subsection{Qualitative Evaluation}
\begin{figure}
    \centering
    \begin{subfigure}[b]{0.3\textwidth}
        \centering
         \includegraphics[width=\textwidth]{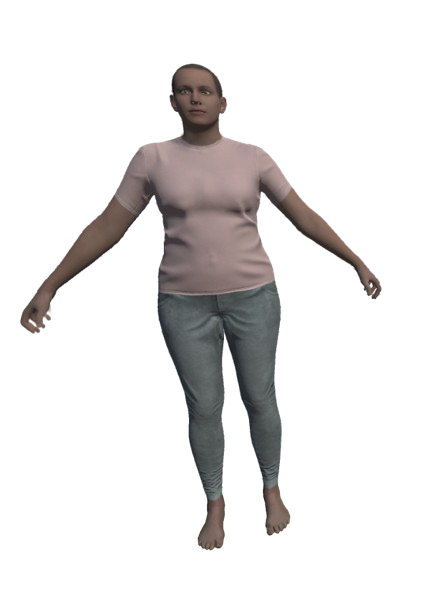}
         \caption*{\centering{"A tall person with very long legs."}}
    \end{subfigure}
    \hfill
    \begin{subfigure}[b]{0.3\textwidth}
        \centering
         \includegraphics[width=\textwidth]{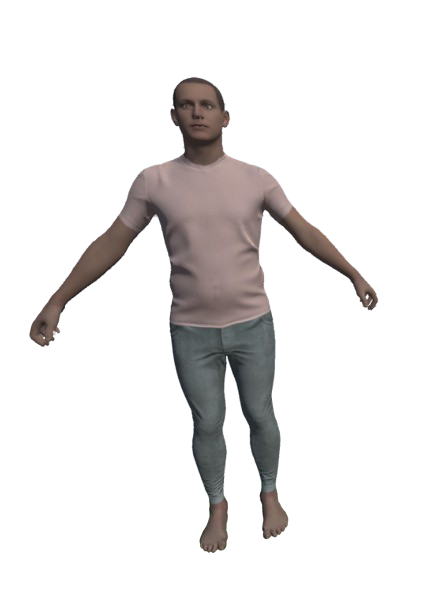}
         \caption*{\centering{"A tall person with very short legs."}}
    \end{subfigure}
    \hfill
    \begin{subfigure}[b]{0.3\textwidth}
        \centering
         \includegraphics[width=\textwidth]{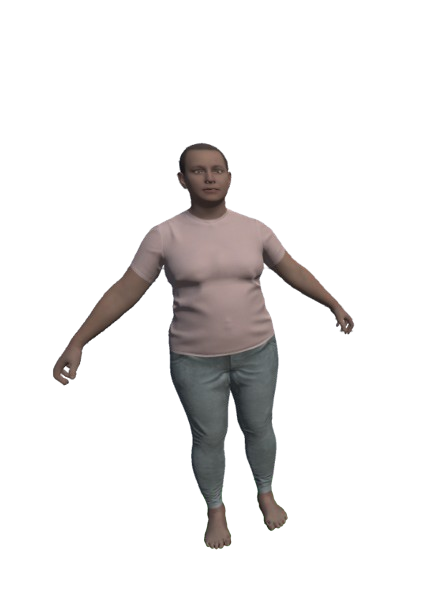}
         \caption*{\centering{"Short, pearl-shaped person."}}
    \end{subfigure}
    \\
    \begin{subfigure}[b]{0.3\textwidth}
        \centering
         \includegraphics[width=\textwidth]{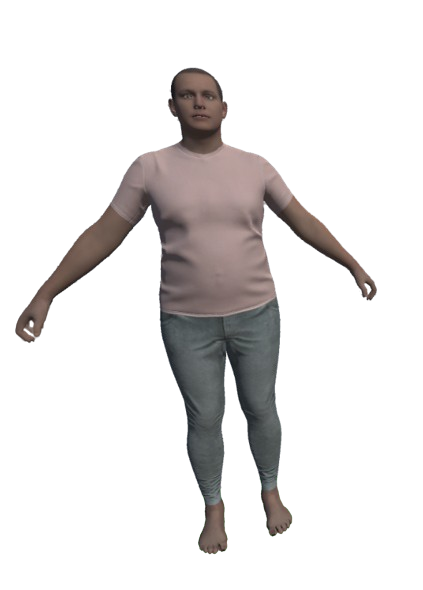}
         \caption*{\centering{"Towering, muscular figure."}}
    \end{subfigure}
    \hfill
    \begin{subfigure}[b]{0.3\textwidth}
        \centering
         \includegraphics[width=\textwidth]{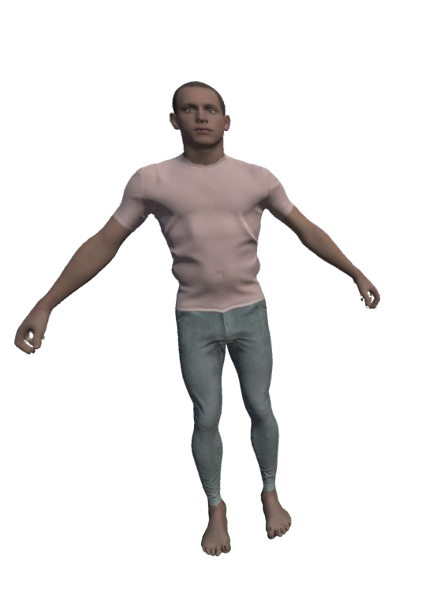}
         \caption*{\centering{"Big shoulders but small hips person."}}
    \end{subfigure}
    \hfill
    \begin{subfigure}[b]{0.3\textwidth}
        \centering
         \includegraphics[width=\textwidth]{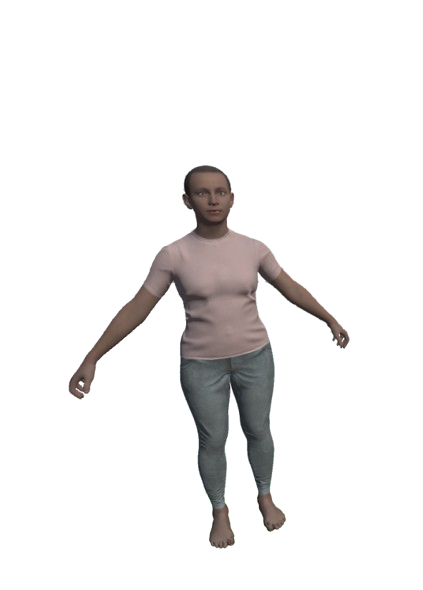}
         \caption*{\centering{"A petite individual that has long arms."}}
    \end{subfigure}
    
    \caption{Generated avatar with different prompts}
    \label{fig:qualitative}
\end{figure}

Figure  \ref{fig:qualitative} showcases some of the avatars generated with the proposed model.
We demonstrate that our model is robust, and has the capability to generate accurate 3D avatars from diverse natural language descriptions, validating the effectiveness of the fine-tuning process and the customized dataset.
%


%

\section{Conclusions}
The paper successfully presented a neural network capable of generating 3D avatars from natural language descriptions. This was achieved through the meticulous creation of a comprehensive dataset, which involved taking measurements of randomly generated avatars, analyzing their distribution, and assigning detailed verbal descriptions. The training process employed fine-tuning techniques such as LoRA and quantization, enabling efficient model training within the constraints of available computational resources. The fine-tuning of large language models (LLMs) specifically for avatar generation has proven highly effective, achieving elevated levels of precision. The ability to generate an avatar from a textual description in just a few seconds opens up numerous practical applications, particularly in fields like storytelling, where dynamic and personalized character creation is highly valued.
\section{Future Work}
Future work could explore further optimization of the network, aiming to enhance performance and accuracy. This includes expanding the dataset to encompass more diverse descriptions, thus improving the model's ability to handle a wider range of inputs. Additionally, integrating pose generation capabilities, potentially in combination with models like PoseGPT or MotionGPT, would enable the creation of complete avatars with both animation and form from a single phrase. This would significantly enhance the dynamism and expressiveness of the generated avatars.

Another important avenue for future research is to train the network to recognize and appropriately respond to descriptions indicating the subject's gender, whether masculine, feminine, or neutral. This involves teaching the network to understand the differences in SMPL-X shape parameters for different genders, ensuring that the generated avatars accurately reflect these distinctions. By addressing these areas, future developments can create more sophisticated and versatile avatar generation systems, broadening their applicability and improving their accuracy and realism.

\begin{credits}

\end{credits}

%
%
%
\bibliographystyle{splncs04}
\bibliography{BodyShapeGPT}

\end{document}